\documentclass{article}
\usepackage{spconf,amsmath,graphicx}
\usepackage{graphicx}
\usepackage{hyperref}
\usepackage{amsfonts,amssymb}
\usepackage{amsmath}
\usepackage{cleveref}
\usepackage{booktabs}
\usepackage{threeparttable}
\usepackage{xcolor}
\usepackage{multirow}
\usepackage{array}

\newcommand{\fin}{\color{black}}
\usepackage{enumitem}
\setlist{nosep, leftmargin=14pt}

\usepackage{mwe} 

\title{
Multi-Modal Active Learning for Automatic Liver Fibrosis Diagnosis based on Ultrasound Shear Wave Elastography
}
\name{Author(s) Name(s)\thanks{Some author footnote.}}
\address{Author Affiliation(s)}

\name{Lufei Gao$^1\dagger$,
	Ruisong Zhou$^2\dagger$\thanks{$\dagger$ Equal contribution.},
	Changfeng Dong$^{3*}$,
	Cheng Feng$^3$,
	Zhen Li$^1$,	
	Xiang Wan$^1$,
	Li Liu$^{1*}$\thanks{* Corresponding author.}
}
\address{
$^1$Shenzhen Research Institute of Big Data, the Chinese University of Hong Kong, Shenzhen, China\\
$^2$Huazhong University of Science and Technology, Wuhan, China\\
$^3$Shenzhen Third People’s Hospital
Shenzhen, China\\
}

\begin{document}
%
\maketitle
\begin{abstract}

With the development of radiomics, noninvasive diagnosis like ultrasound (US) imaging plays a very important role in automatic liver fibrosis diagnosis (ALFD).
Due to the noisy data, expensive annotations of US images, the application of Artificial Intelligence (AI) assisting approaches encounters a bottleneck. Besides, the use of mono-modal US data limits the further improve of the classification results. In this work, we innovatively propose a multi-modal fusion network with active learning (MMFN-AL) for ALFD to exploit the information of multiple modalities, eliminate the noisy data and reduce the annotation cost.
Four image modalities including US and three types of shear wave elastography (SWEs) are exploited. A new dataset containing these modalities from 214 candidates is well-collected and pre-processed, with the labels obtained from the liver biopsy results. Experimental results show that our proposed method outperforms the state-of-the-art performance using less than 30\% data, and by using only around 80\% data, the proposed fusion network achieves high AUC 89.27\% and accuracy 70.59\%.

\fin
\end{abstract}
\begin{keywords}
Liver fibrosis diagnosis, Shear wave elastography, Active learning, Attention, Multi-modal fusion
\end{keywords}
\section{Introduction}
\label{sec:intro}

Liver fibrosis is a main prognostic factor in chronic liver disease (CLD) patients' treatment \cite{vergniol2011noninvasive}.
To assess the severity of liver fibrosis \cite{kose2015evaluation} clinically, liver biopsy (LB) is employed as the gold standard to classify liver fibrosis into 5 stages under the METAVIR scoring system: no fibrosis (F0), mild fibrosis (F1), significant fibrosis (F2), severe fibrosis (F3) and cirrhosis (F4) \cite{pol2014non}.
However, LB is an invasive procedure with potential complications.
To overcome its limitations, serum indexes and multiple imaging modalities, such as B-mode ultrasound (US), computed tomography (CT), magnetic resonance imaging (MRI), magnetic resonance elastography (MRE) and US-based elastography are used for noninvasive automatic liver fibrosis diagnosis (ALFD).
However, using these techniques requires specialized equipment and well-trained professionals while there is a paucity of these resources in regions CLD is prevalent.
Therefore, automatic and cost-effective AI based methods are supposed to be developed to improve the monitoring and treatment of CLD patients worldwide.

	
	
In the literature, some works \cite{choi2018development,yasaka2018deep, yasaka2018liver} using CT and MRI for automatic assessment and
some employ simple US images which are more convenient, harmless and cheaper \cite{khvostikov2017ultrasound}.
However, all above systems fail to discriminate early stages of liver fibrosis.	
In contrast, using US-based elastography images exhibits great potential at assessing all stages of liver fibrosis.
It can be performed with different techniques, including transient, point (STQ), and 2D-SWE (STE) \cite{sigrist2017ultrasound}.
Recently, several methods were designed based on STE. For instance,
Wang et al. \cite{wang2019deep} constructed a four-layer convolutional network. A fully connected layer was used as a two-class classifier (F4 vs. F0-F3; F3-F4 vs. F0-F2; F2-F4 vs. F0-F1). In \cite{xue2020transfer}, a transfer learning radiomics model combining the information from grayscale and elastography images was proposed for liver fibrosis grading, achieving AUCs 95$\%$, 93.2$\%$, and 93$\%$ for classifying F4, $\geq$ F3, and $\geq$ F2, respectively. They are the state-of-the-art (SOTA) performance as far as we know.

However, there are several limitations with the previous methods. Firstly, they classified liver fibrosis into less than or equal to four stages, since two neighboring stages are hard to be discriminated.
Secondly, in clinical workflow, the doctor’s diagnosis is usually based on more than one type of medical images or indicators rather than a single modality. Thirdly, the AI based methods suffer from the limited annotated noisy data, since it is expensive and tedious to collect, clean and annotate numerous images.

Active Learning (AL) has been successfully deployed into many applications \cite{sun2015active,beluch2018power,liu2017active}, which indicates that AL is a good choice
for reducing the annotation cost, removing the noisy data automatically, and improving the robustness of the model. There have been many
popular selection strategies in the literature, mainly including
query by committee, expected error reduction and uncertainty sampling.

	
	To this end, we aim to investigate the effectiveness of different modalities in deep neural network with AL, which aims to achieve effective performance given a limited amount of annotated data selected from a noisy data pool.
	A novel \textit{multi-modal fusion network with AL (MMFN-AL)} to assess the degree of liver fibrosis is proposed.
	Specifically, we implement the AL to select the informative samples from the unlabeled dataset, and these samples are then used to fine-tune the MMFN, i.e. a pre-trained ResNet50 followed squeeze-excitation (SE) attention modules \cite{hu2018squeeze}.
	We first implement a mono-modal network as the baseline by transfer learning with ResNet-50 \cite{he2016deep}. Then a feature-level fusion network with SE blocks is designed to effectively integrate complementary information from different modalities.
	Finally, the mono-modal and bi-modal networks with different modality combinations are evaluated.
	
\begin{figure}
		\includegraphics[width=0.5\textwidth]{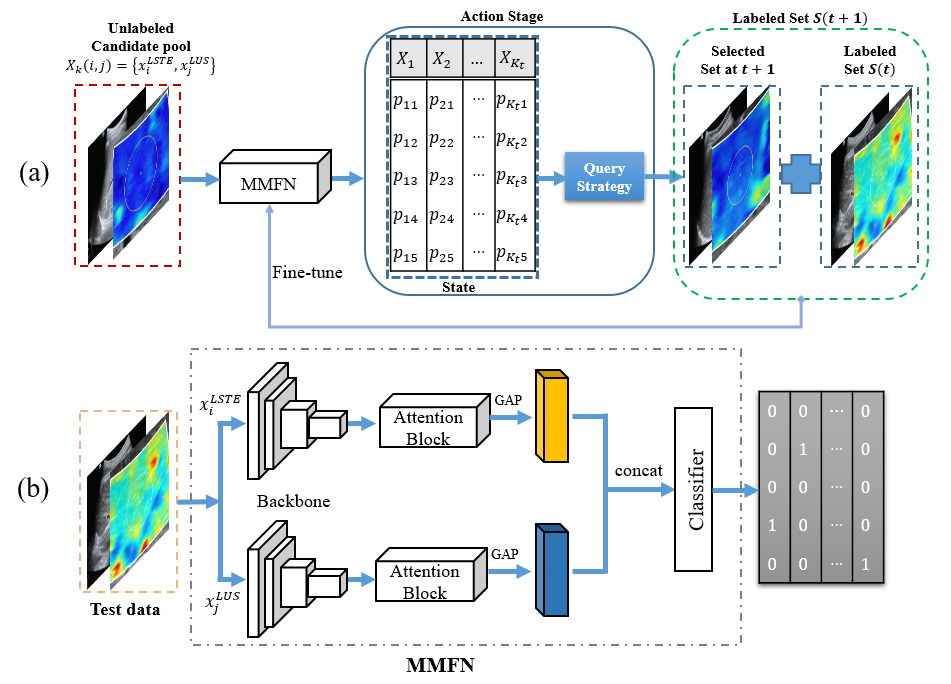}
		\caption{The architecture of the proposed MMFN-AL, where (a) is the AL stream for training and fine-tuning MMFN and (b) is the MMFN framework for the ALFD task. The backbone structure is Resnet-50.}
		\label{fig:multimodal1}
	\end{figure}

	Experimental results demonstrate our hypothesis that integrating multiple modalities into the network significantly improves the diagnosis accuracy and outperform the SOTA. Furthermore, the use of AL can eliminate certain noisy samples and further improve the performance using less than 30\% labelled data.
	An overview of our proposed model is shown in Fig. \ref{fig:multimodal1}.
	To conclude, this paper contains the following contributions:
	\begin{itemize}
		\item  To the best of our knowledge, this is the first work that build a novel large-scale US dataset containing four medical image modalities of liver for ALFD.
		\item A novel multi-modal fusion network with AL is proposed, effectively reducing labelling efforts, eliminating the noisy data and improving the robustness of the model.
		\item Experimental results show that our MMFN-AL achieves better AUCs with less than 30\% data than the SOTA methods with 100\% data.
	\end{itemize}

\section{Multi-Modal Dataset Construction}
A dataset consisting of four modalities is constructed in this work.
They were collected in the clinical examination of liver fibrosis, named by liver STE images (LSTE), spleen STE images (SSTE), liver STQ images (LSTQ) and liver ultrasound images (LUS), as illustrated in Fig. \ref{fig:dataexample}.

\subsection{Clinical Flow}
The data were collected from a hospital by one radiologist who was strictly trained for ultrasound examination with SWE measurement using the uniform procedure.
The routine examination of a CLD patient includes blood sampling, physical examination, US examination and SWE measurement. The corresponding five-stage METAVIR score was given by the result of LB.

\begin{figure}
\includegraphics[width=0.5\textwidth]{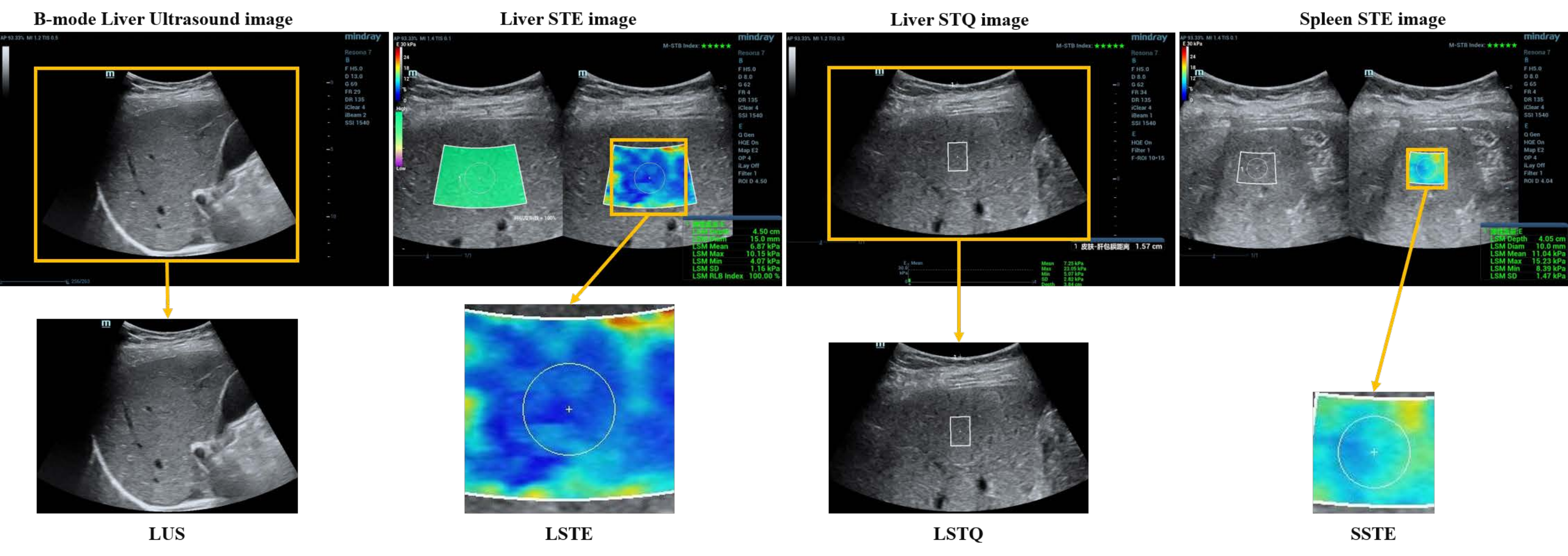}
\caption{Dataset illustration.}
\label{fig:dataexample}
\end{figure}

In the clinical examination of one patient, the doctor collected seven US images, five LSTE images, three to five LSTQ images and three to five SSTE images that reflect liver fibrosis to a certain extent.
In addition, the doctor collected ultrasound videos from some patients. By video sampling, we obtain 20\-30 more LUS images for each patient with video data.
The illustration of these four kinds of images was shown in the first line of Fig. \ref{fig:dataexample}.
The regions inside the yellow box are taken as the final inputs.

\vspace{-0.5cm}
\subsection{Multi-Modal Dataset}
We totally collect the samples for 214 candidates from a hospital, containing 213 LSTE candidates, 191 LUS candidates, 201 LSTQ candidates and 190 SSTE candidates. It should be noted that patients could have several modalities, but not necessary all modalities.
To make the evaluation fair and reasonable, only the patients with four modalities were included into our final dataset, resulting in 168 such patients, where 41 are F0, 51 are F1, 31 are F2, 27 are F3 and 18 are F4.
To augment the data pairs in the bi-modal experiments, we carry out pairwise combinations between every two modalities. A data summary of image amounts is listed in Table \ref{tab:DataSummary}.

\begin{table}[!h]\scriptsize
\caption{The number of images in the collected set for each data modal with five liver fibrosis stages.}
\label{tab:DataSummary}
\centering\begin{tabular}{c||p{0.45cm}<{\centering}|p{0.45cm}<{\centering}|p{0.45cm}<{\centering}|p{1.1cm}<{\centering}|p{1.2cm}<{\centering}|p{1.2cm}<{\centering}} \hline
     Modal & LSTE & LUS   & LSTQ  & LSTE$\times$LUS & LSTE$\times$LSTQ  & LSTE$\times$SSTE \\ \hline \hline
Total & 778  & 3572  & 683   & 16592      & 3232     & 3171    \\ \hline
F0    & 185  & 933   & 183   & 4195       & 859      & 827 \\ \hline
F1    & 235  & 1095  & 204   & 5077       & 958      & 938 \\ \hline
F2    & 142  & 604   & 117   & 2818       & 548      & 548 \\ \hline
F3    & 132  & 515   & 109   & 2509       & 535      & 526 \\ \hline
F4    & 84   & 425   & 70    & 1993       & 332      & 332 \\ \hline
\end{tabular}
\end{table}

\vspace{-0.3cm}
\section{Methodology}
\subsection{Feature Extraction Networks}
Considering the multi-modal inputs, we design the MMFN\-AL as illustrated in Fig. \ref{fig:multimodal1}. Each stream consists of a feature extractor module and an attention block.
To extract deeper and handle high-level features from medical images, we exploit the idea of the residual and adapt ResNet-50 \cite{he2016deep} that was pretrained on
ImageNet \cite{deng2009imagenet} as the feature extractor.
Rather than the multi-stream network used in video action recognition \cite{feichtenhofer2016convolutional} for spatially correlated data streams,
we consider spatially uncorrelated characteristics of the multi-modal images in this task, and design our model to perform the fusion after
the global average pooling (GAP) layer, which removes the spatial information by averaging each feature map into a single value.

\subsection{Multi-Modal Active Learning Framework}

\subsubsection{Multi-Modal Fusion}
For multi-modal inputs, we believe that the importance of each modal is different.
With the goal of improving the feature quality of representations produced by each stream, we add a down-sampling layer and an SE attention block after feature extraction.

For each stream input $I \in \mathbb{R}^{c \times h \times w}$, where $c, h$ and $w$ denote the number of channels, height and width of the image, respectively, we convert the medical images from other format to RGB. Let $F \in \mathbb{R}^{c' \times \frac{h}{s} \times \frac{w}{s}}$ be the array of feature maps generated by the feature extractor in each branch.
The value of channel numbers $c'$ and down-sampling rate $s$ depend on the backbone we used, which are 2048 and 32 in ResNet-50. In order to reduce the redundant features, we exploit a point-wise convolution to reduce the dimensions to 256.
Then we use an SE module to enhance the feature representations after feature extraction by the ResNet backbone.
We aggregate the spatial information of each feature map by average pooling to generate a spatial feature $\mathbf{F_{avg}^{c}}$. This spatial feature is forwarded to a multi-layer perceptron (MLP) ultimately to produce our channel attention map $\mathbf{M_c} \in \mathbb{R}^{c' \times 1 \times 1}$,
and merge the output feature vectors using element-wise summation. In short, the attention block can be computed as:
\begin{gather*}
  \mathbf{M_c(F)} = \sigma(MLP(AvgPool(\mathbf{F}))) = \sigma(\mathbf{W_1}(\mathbf{W_0}(\mathbf{F_{avg}^{c}}))),  \\
  \mathbf{M_o(F)} = \mathbf{M_c(F)}\otimes \mathbf{F},
\end{gather*}
where $\sigma$ denotes the sigmoid function, $\mathbf{W_0}$ and $\mathbf{W_1}$ are the MLPs weights. Note that the ReLU activation function is followed by $\mathbf{W_0}$.
And $\otimes$ denotes element-wise multiplication.

Our fusion layer is implemented by separately feeding $\mathbf{M_o(F)}$ into a GAP layer to obtain several 256-dimensional vectors corresponding to different modalities, which are then concatenated to form a $ 256\times n$ vector that contains information from $n$ modalities.
For classification, the combined vector is fed into a fully connected (FC) layer to produce posterior probability for each class, which is predicted by selecting the class with the maximum posterior probability.


\subsubsection{Query Strategy for Active Learning}
At each AL iteration $t$, the action stage in Fig.\ref{fig:multimodal1}(a) is to select unlabeled images from the candidate pool in order to form a new labeled set to fine-tune the MMFN model.
Let $S^{L}(t)$ denote the index set of modalily $L\in\{\text{LSTE}, \text{LUS}, \text{LSTQ}, \text{SSTE}\}$ at iteration $t$.
For instance, taking LSTE and LUS as the input modalities, the unlabeled candidate pool is denoted as $\mathbf{X}(t)=\{X_k(i,j)\}$, where $(i,j)\in S^{\text{LSTE}}(t) \times S^{\text{LUS}}(t)$.
Specifically, $X_k(i,j)=\{(x^{\text{LSTE}}_i, x^{\text{LUS}}_j)\}$, where $k = 1,2, ..., K_t^{(\text{LSTE}, \text{LUS})}$.
Here, $K_t^{(\text{LSTE}, \text{LUS})}=\Pi_{L\in\{\text{LSTE}, \text{LUS}\}}|S^{L}(t)|$ is the total amount of unlabeled data.

After one training iteration, the MMFN provides a prediction state $P_k=[p_{k1}, ..., p_{k5}]$ for each input $X_k(i,j)$.
If $(i,j)$ is selected as the labeled data according to $P_k$ and the query strategy, $X_k(i,j)$ will be removed from the candidate pool $\mathbf{X}(t)$.
Random sampling strategy (RAND) abd the entropy sampling strategy (ES) are employed in this work for AL.

Information entropy of an input $X_k$ is calculated on the output probability of the classifier which is trained at the previous iteration. It is denoted as:
$
H(X_k) =-\sum_{l=1}^L p(Y_k=l\mid X_k)\log p(Y_k=l\mid X_k),
$
where $p(Y_k=l\mid X_k)$ is the probability that the input image $X_k$ is predicted as label $j$.
Higher entropy value means that selected samples carry richer information.
In the implementation, we take the average probability of $n$ prediction states to calculate the entropy for each input data and select $n_{query}$ data with the largest entropy values, named entropy sampling with dropout (ESD).

\section{Experiments}
\subsection{Implementation Details}
We first implemented mono-modal experiments using the ResNet-50 based neural network to investigate how each modal is performed.
The input images of each modal were illustrated in the second row of Fig. \ref{fig:dataexample}.
For LSTQ and LUS, the complete LUS image was selected. For LSTE (or SSTE), the square with the size of 224$\times$224 pixels including the entire ROI was selected as the input.
For evaluation, $80\%$ of the patients in each stage was selected to constitute the training dataset and the rest for testing.
	
To compare with the SOTA, we reproduced two previous methods, named by DLRE \cite{wang2019deep} and IncepV3 \cite{xue2020transfer} on our dataset.
For a fair comparison, we replace the two-class or four-class classifier in the original models with a five-class FC classifier.
	
\subsection{Evaluation Metrics}
The area under the receiver operating characteristic curve (AUC) was used as the main index for performance evaluation. The average AUC for the five-category classification is used as an overall comparison.
Accuracy (Acc) is also computed to evaluate the performance.

\subsection{Results and Discussions}
Diagnostic accuracies and five-class average AUCs of all mono-modal networks for three modalites are shown in Table 2 and Table 3. SSTE is not used for mono-modal network evaluation because it is not from the liver, mismatched with our task target.
ResNet50-RAND and ResNet50-ESD represents the corresponding query strategies employed by AL.
The proposed fusion model MMFN is evaluated on the bi-modal cases, i.e. fusing LSTE with LUS, LSTQ and SSTE, respectively.
The effectiveness of combining MMFN with AL is further evaluated using RAND and ESD.
The results relative with network fusion are illustrated in Table 4.

\vspace{-0.4cm}
\begin{table}[ht]\scriptsize
\label{tab:compare1}
\caption{Accuracy and average AUC of mono-modal for different modalities. }
\centering
\begin{tabular}{c||c|c|c|c|c|c}
\hline
\multirow{2}{*}{Score (\%)} & \multicolumn{2}{c|}{DLRE\cite{wang2019deep}} & \multicolumn{2}{c|}{IncepV3\cite{xue2020transfer}} & \multicolumn{2}{c}{ResNet50} \\ \cline{2-7}
 & \textit{Acc} & \textit{AUC} & \textit{Acc} & \textit{AUC} & \textit{Acc} & \textit{AUC} \\ \hline
\textbf{LSTE} & \textbf{61.76} & 84.19 & 47.06 & 83.74 & 55.88 & \textbf{84.43} \\ \hline
LUS & 32.35 & 62.67 & \textbf{41.18} & \textbf{77.34} & 32.35 & 74.94  \\ \hline
LSTQ & 38.24 & 63.62 & 38.24 & 71.93 & \textbf{47.06} & \textbf{75.87}  \\ \hline
\end{tabular}
\end{table}
\vspace{-0.4cm}
\begin{table}[ht]\scriptsize
\label{tab:mono_al}
\caption{Results of mono-modal network with active learning. (d): percentage of the labelled data.}
\centering
\begin{tabular}{c||c|c|c|c}
\hline
\multirow{2}{*}{Score (\%)} & \multicolumn{2}{c|}{ResNet50-RAND} & \multicolumn{2}{c}{\textbf{ResNet50-ESD}} \\ \cline{2-5}
& \textit{Acc} & \textit{AUC (d)} & \textit{Acc} & \textit{AUC(d)} \\ \hline
LSTE & 64.71 & 87.85 (26.9\%) & \textbf{65.12} & \textbf{89.72(87.4\%)} \\ \hline
LUS  & 35.29 & 68.75(65.0\%) & 44.12 & 72.95(32.3\%) \\ \hline
LSTQ & 47.6 & 72.6 (39.6\%) & 52.94 & 72.86(89.2\%) \\ \hline
\end{tabular}
\end{table}

\vspace{-0.4cm}
\begin{table}[h]\scriptsize
\label{tab:compare2}
\caption{Accuracy and average AUC in \% of MMFN with different modal combinations. $\oplus$: operation of concatenation of feature maps. (d): percentage of the labelled data.}
\centering
\begin{tabular}{c||c|c|c|c|c|c}
\hline
\multirow{2}{*}{} & \multicolumn{2}{c|}{\textbf{MMFN}} & \multicolumn{2}{c|}{\textbf{MMFN-RAND}} & \multicolumn{2}{c}{\textbf{MMFN-ESD}} \\ \cline{2-7}
LSTE$\oplus$ & \textit{Acc} & \textit{AUC} & \textit{Acc} & \textit{AUC (d)} & \textit{Acc} & \textit{AUC (d)} \\ \hline
LUS & 58.82 & 88.84 & 61.76 & 86.78 (12.5\%) & 64.71 & 88.08 (60.0\%) \\ \hline
\textbf{LSTQ} & \textbf{64.71} & \textbf{87.50} & \textbf{70.59} & \textbf{89.12 (79.5\%)} & \textbf{64.71} & \textbf{89.27 (79.5\%)} \\ \hline
SSTE & 61.76 & 84.13 & 67.65 & 84.58 (79.7\%) & 67.65 &  85.03 (54.8\%)\\ \hline
\end{tabular}
\end{table}

\subsubsection{Comparison with the SOTA}
As shown in the first six columns of Table 2, the best average AUCs of the mono-modal models using LSTE and LSTQ are achieved by the ResNet50 backbone.
By comparing the SOTA results with the results in Table 3 and Table 4, it illustrates that our proposed MMFN-AL architecture outperforms the two previous methods.

\vspace{-0.4cm}
\subsubsection{Effectiveness of multi-modal fusion network}
We train ResNet50-based transfer learning network as the baseline for our experiment.
As shown in Table 2, LSTE image presents the best performance among three mono-modal experiments, among which DLRE provides the best diagnosis accuracy while ResNet50 gives the best AUC.
This demonstrates that LSTE images carry more effective information for liver fibrosis grading.
By fusing any other modality with LSTE, the performances are all improved significantly, and the fusion of LSTE and LSTQ gives the best diagnosis accuracy 70.59\% and best average AUC 89.27\% among the bi-modal experiments.
The results justify the advantage of fusing multiple informative modalities in the DL systems for automatic diagnosis.

\vspace{-0.4cm}
\subsubsection{Effectiveness of active learning}

Comparing the last two columns of Table 2 with Table 3, it shows that the LSTE mono-modal case most benefits from the AL.
Both accuracy and AUC are improved significantly and it achieves a relatively high AUC with only 26.9\% data.
In terms of accuracy, the case of LSTQ also performs much better by applying AL. The accuracy is increased by more than five percentage using 89.2\% data.
For multi-modal cases, the effectiveness of AL is obvious as well.
Specifcally, the best diagnosis accuracy among all experiments is achieved by LSTE$\oplus$LSTQ using random AL strategy, by taking 79.5\% data from the candidate pool.
For LSTE$\oplus$LUS, by taking only 12.5\% data, it can achieve relatively high accuracy and AUC values.

\section{Conclusion}
In this work, a multi-modal fusion network with active learning is innovatively proposed for ALFD.
As far as we know, it is the first time that a five-class classification model is developed and a dataset consisting of four modalities is constructed for this task.
The proposed models are compared with two SOTA methods in the mono-modal case and demonstrate relatively better performances using different image modalities.
Our experiment indicates that the MMFN-AL with the combination of two proper modalities reaches the best average AUC.
In the future, query strategies are expected to be designed for the specific task, and the ability for accurately discerning stage F1 and F2 needs to be improved according to the clinical requirements.
\fin

\section{Compliance with Ethical Standards}
\label{sec:ethics}

We claim that we do not have any compliance for this work.

\section{Acknowledgments}
\label{sec:acknowledgments}

We would like to thanks the hospital to offer the data, and the doctors for data collection and annotation.

\bibliographystyle{IEEEbib}

\end{document}